\pdfoutput=1
\documentclass[11pt]{article}

\usepackage[]{EMNLP2023}

\usepackage{times}
\usepackage{latexsym}

\usepackage[T1]{fontenc}
\usepackage[utf8]{inputenc}

\usepackage{microtype}

\usepackage{inconsolata}
\usepackage{amsmath}
\usepackage{algorithmicx}
\usepackage{algorithm}
\usepackage{amssymb}
\usepackage{graphicx}
\usepackage{subfigure}
\usepackage{multirow}
\usepackage{booktabs}
\usepackage{bbm}
\usepackage{bm}
\usepackage{algpseudocode}
\usepackage{bbding}
\usepackage{pifont}
\usepackage{color}
\newcommand{\cmark}{\ding{51}}%
\newcommand{\xmark}{\ding{55}}%

\usepackage{xcolor}
\definecolor{green(ncs)}{rgb}{0.0, 0.62, 0.42}
\definecolor{frenchblue}{rgb}{0.0, 0.45, 0.73}

\newcommand{\ie}{\textit{i}.\textit{e}.}
\newcommand{\eg}{\textit{e}.\textit{g}.}

\title{Beneath the Surface: Unveiling Harmful Memes with  Multimodal Reasoning Distilled from Large Language Models}

\author{Hongzhan Lin$^{1}$, Ziyang Luo$^{1}$, Jing Ma$^{1}$\thanks{\; Jing Ma is the corresponding author. The first two authors contributed equally to this work.}, Long Chen$^2$ \\
        $^1$Hong Kong Baptist University \\ $^2$The Hong Kong University of Science and Technology \\ 
        \texttt{\{cshzlin,cszyluo,majing\}@comp.hkbu.edu.hk},
        \texttt{longchen@ust.hk}}

\begin{document}
\maketitle

\begin{abstract}
The age of social media is rife with memes. Understanding and detecting harmful memes pose a significant challenge due to their implicit meaning that is not explicitly conveyed through the surface text and image. However, existing harmful meme detection approaches only recognize superficial harm-indicative signals in an end-to-end classification manner but ignore in-depth cognition of the meme text and image. In this paper, we attempt to detect harmful memes based on advanced reasoning over the interplay of multimodal information in memes. Inspired by the success of Large Language Models (LLMs) on complex reasoning, we first conduct abductive reasoning with LLMs. Then we propose a novel generative framework to learn reasonable thoughts from LLMs for better multimodal fusion and lightweight fine-tuning, which consists of two training stages: 1) Distill multimodal reasoning knowledge from LLMs; and 2) Fine-tune the generative framework to infer harmfulness. Extensive experiments conducted on three meme datasets demonstrate that our proposed approach achieves superior performance than state-of-the-art methods on the harmful meme detection task. 
\end{abstract}

\section{Introduction}
The development of social media platforms has given rise to a new form of multimodal content known as: \textbf{meme}. A meme typically comprises a picture that is combined with a concise text component.
Memes possess the capacity to quickly spread across the internet, especially on social media platforms, due to their ease of dissemination. %While memes are commonly perceived as humorous, they have the potential to become a source of harm when the combination of images and text is ingeniously employed to exploit political and sociocultural divisions. 
While memes are often seen as humorous, there is a potential for harm when the combination of images and texts is strategically used to promote political and sociocultural divisions. For instance as in Figure~\ref{fig:harmful_}, during the COVID-19 pandemic, a widely circulated meme falsely claimed that the mRNA vaccine would alter human genetic code (DNA)\footnote{\url{https://www.bbc.com/news/55101238}}.
%a popular meme shown in Figure~\ref{meme_illustration} spread disinformation that a vaccine will modify our genetic code (DNA)\footnote{\url{https://www.bbc.com/news/55101238}}.
%Such multimodal disinformation fed into concerns about the efficacy or safety of vaccines and harmed the achievements of epidemic prevention in related countries or regions of the world. 
% The dissemination of such multimodal disinformation fueled anxieties regarding vaccine effectiveness and safety, impeding the establishment of robust immune barriers in the affected countries or regions worldwide
Such multimodal disinformation spread caused vaccine safety and effectiveness concerns, hindering the formation of strong immune defenses in impacted areas globally~\cite{ basch2021global, lin2022detect}. Besides, another meme example shown in Figure~\ref{fig:harmful} perpetuates harmful stereotypes and generalizations about Asians.
% \lc{the description of this example is verbose and over-long.} 
Therefore, it is necessary to develop automatic approaches to facilitate harmful meme detection for unveiling the dark side of memes.

% \begin{figure}[t]
% \centering
% \scalebox{0.28}{\includegraphics{task_illustration.pdf}}
% \caption{Example of trending memes on social media. \lc{Suggestion: you can merge current Fig 1 and Fig 2. In this new Figure 1, you can show more examples of memes, some are harmful, and some are harmless.}}
% \label{meme_illustration}
% \vspace{-0.5cm}
% \end{figure}

Harmful memes\footnote{\textcolor{red}{\textbf{Disclaimer}: \textit{This paper contains discriminatory content that may be disturbing to some readers, where meme examples and words are offensive or hateful in nature. These contents are provided for illustrative purposes only and do not represent the views and standpoints of the authors.}}} are generally defined as ``multimodal units consisting of an image and accompanying text that has the potential to cause harm to an individual, an organization, a community, or the whole society”~\cite{sharma2022detecting}.
% \footnote{\color{red}\textbf{Disclaimer:} This paper contains content that may be disturbing to some readers.}. 
%Despite the considerable influence of memes, their multimodal nature and covert meaning render them highly challenging to comprehend and analyze.
%Generally, understanding and analyzing memes poses a significant challenge due to their implicit meaning that is not explicitly conveyed through the surface text and image. 
Previous studies~\cite{kiela2020hateful, pramanick2021detecting, pramanick2021momenta} attempted to straightforwardly utilize pre-trained vision-language models~\cite{li2019visualbert, lu2019vilbert} for harmful meme detection by training additional task-specific classification layers. More recently, \citet{cao2023prompting} proposed a prompt-tuning method with the meme text and image caption as the prompt for masked language modeling~\cite{devlin2018bert, liu2019roberta}.

{
\begin{figure*}[t]
\centering
\subfigure[Harmful]{
\begin{minipage}[t]{0.3\linewidth}
\centering
\scalebox{0.95}{\includegraphics[width=5cm]{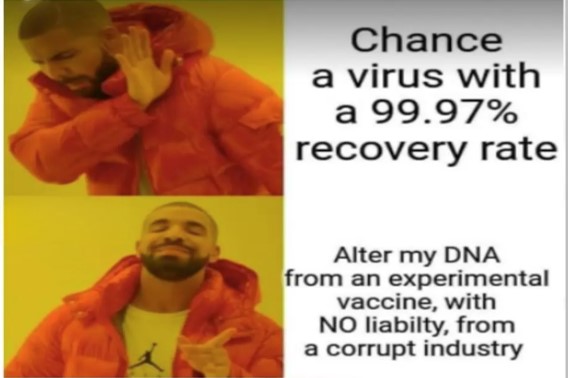}}
%\caption{fig2}
\label{fig:harmful_}
\end{minipage}%
}%
\subfigure[Harmful]{
\begin{minipage}[t]{0.3\linewidth}
\centering
\scalebox{0.95}{\includegraphics[width=5cm]{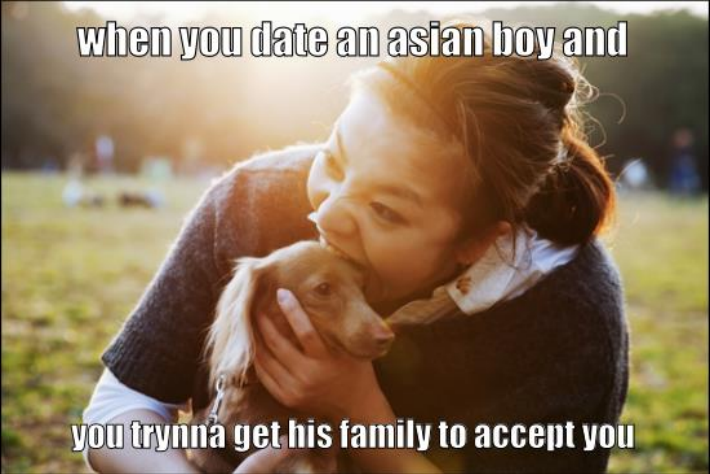}}
%\caption{fig1}
\label{fig:harmful}
\end{minipage}%
}%
\subfigure[Harmless]{
\begin{minipage}[t]{0.3\linewidth}
\centering
\scalebox{0.95}{\includegraphics[width=5cm]{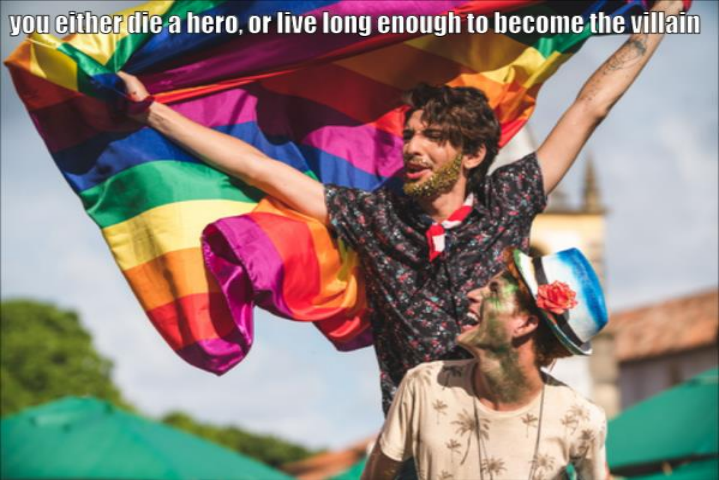}}
%\caption{fig2}
\label{fig:harmless}
\end{minipage}%
}%
\centering
\vspace{-0.3cm}
\caption{Examples of harmful and harmless memes. \textbf{Meme text}: (\textbf{a}) \textit{Chance a virus with a 99.97\% recovery rate; Alter my DNA from an experimental vaccine, with NO liability, from a corrupt industry.} (\textbf{b}) \textit{when you date an asian boy and you trynna get his family to accept you.} (\textbf{c}) \textit{you either die a hero, or live long enough to become the villain.}}
\label{fig:motivation}
\vspace{-0.2cm}
\end{figure*}}

However, existing harmful meme detection approaches oversimplified the problem as an end-to-end classification paradigm, which only recognizes the superficial signals conveyed through the surface text and image. But more in-depth investigation and cognition on the implicit meaning is required especially when the image and text are not obviously correlated~\cite{pramanick2021momenta}. Intuitively, the key to harmful meme detection is to excavate rich correlations beneath the surface of the seemly uncorrelated text and image in the meme: 1) For example as in Figure~\ref{fig:harmful}, the image and the text are not harmful when considered in isolation, but are harmful when taken as a whole. A human checker should cognize that, the ``biting'' action in the image of a young woman with her pet dog ridicules Asians’ ``dog-eating'' behavior, which corresponds to the ``asian'' word in the text. 2) In contrast, some harmful signals (\eg, ``die'' or ``villain'') are observed in the text of Figure~\ref{fig:harmless}, but the meme itself actually does not promote hate or discrimination against a particular group of people. Because the text is a quote from a popular movie and is often used as a philosophical statement about the choices people make in life. And the image further adds a celebratory and joyful tone to the overall message. In comparison, conventional detection methods just focused on recognizing shallow harm-indicative signals without such multimodal reasoning and essential background knowledge consideration, so the social dynamics of different races or the origin of the meme text from classical movie lines may not be well-cognized. Unlike such recognition-level detection models, we argue that establishing reasonable thought between textual and visual information can further improve meme understanding with background knowledge for better harmful meme detection.

Inspired by the success of LLMs for reasoning at the cognition level with contextual background knowledge~\cite{weichain, kojima2022large, zhang2022automatic}, we propose a novel approach: \textbf{\textsc{Mr.Harm}}, by leveraging the \underline{\textbf{M}}ultimodal \underline{\textbf{r}}easoning knowledge distilled from LLMs for \underline{\textbf{Harm}}ful meme detection. To this end, we first prompt LLMs for abductive reasoning, and then propose a two-stage generative framework based on smaller language models to learn reasonable thoughts from LLMs for better multimodal fusion and lightweight fine-tuning.
% \lc{a more natural order here is: 1) we first prompt LLMs for abductive reasoning, 2) then we use a two-stage generative framework to learn from them.}
%
%We argue that the key insight to the harmful meme detection task that requires seamless integration between recognition and cognition, is to reason and evolve with the cognition-level rationale beyond the recognition-level perception~\cite{davis2015commonsense}. Recent studies~\cite{weichain, kojima2022large, zhang2022automatic} observe one intriguing emerging property of LLMs: the ability to reason at the cognition level with contextual background knowledge in LLMs, which allows it to better capture the inter-relationship between visual and textual elements in memes for accurately detecting harmful memes.
% \textcolor{red}{Inspired by LLM, here requires some justification: why LLM can help? why it can infer all possible correlations between text and image, etc.}
%
%
%To this end, inspired by the powerful expressive ability of LLMs, we propose to detect harmful memes with Multimodal Reasoning knowledge distilled from LLMs, \textsc{Mr.}\textsc{Harm}, which
More specifically, we incorporate the meme text and image into a \textbf{two-stage} training paradigm: 1) \emph{Reasoning Distillation}: In the first stage, we fine-tune our smaller language models with the interaction of language and vision features to distill multimodal reasoning knowledge from LLMs, which empowers our framework with the ability to conduct cognitive reasoning for the harmfulness prediction. 2) \emph{Harmfulness Inference}: In the second stage, we exploit the fine-tuned small language models to infer the final harmfulness prediction. In this manner, we augment the harmful meme detection model with multimodal reasoning knowledge to unmask the implicit meaning hidden in holistic multimodal information from memes.
% \textcolor{red}{so the above method part require to be re-organized, \eg, following a pipeline order.} 

We evaluate our proposed approach based on three public meme datasets. The results not only show that our method outperforms strong harmful meme detection baselines by a large margin, %and Experimental results not only reveal the effectiveness of our proposed framework 
but also provide fine-grained analysis for interpreting how our approach works. Our contributions are summarized as follows in three folds:
\begin{itemize}
\item To our best knowledge, we are the first to alleviate the issue of superficial understanding for harmful meme detection by explicitly utilizing commonsense knowledge, from a fresh perspective on harnessing advanced LLMs.\footnote{Our code is available at \url{https://github.com/HKBUNLP/Mr.Harm-EMNLP2023}}
% To our best knowledge, we are the first to study multimodal reasoning for harmful meme detection from a fresh perspective on harnessing advanced LLMs, which aims to excavate the implicit meaning that is not explicitly conveyed through the surface of the meme text and image. 
% \lc{instead of just saying you are the first one using LLM for this task, it would be better to say you are the first one: finding the weakness of superficial understanding of existing methods, and the first model with EXPLICITLY utilizing commonsense knowledge. After this, LLM is a suitable tool for realizing this goal. } 
%\vspace{-0.21cm}
\item We propose a novel generative framework to fine-tune smaller language models augmented with the multimodal reasoning knowledge distilled from LLMs, which facilitates better multimodal fusion and lightweight fine-tuning for harmfulness prediction.
% \lc{highlight more about the advantage of your current designs: the smaller LM design, two-stage pipeline, and so on.}
%\vspace{-0.21cm}
\item Extensive ablations on three meme datasets confirm that our method could yield superior performance than state-of-the-art baselines for the harmful meme detection task.
\end{itemize}

\section{Related Work}

\subsection{Harmful Meme Detection}
Harmful meme detection is a rapidly growing area in the research community, driven by the recent availability of large meme benchmarks~\cite{kiela2019supervised, suryawanshi2020multimodal, pramanick2021detecting}. The Hateful Memes Challenge organized by Facebook~\cite{kiela2020hateful} further encouraged researchers to develop solutions for detecting harmful memes in hate speech~\cite{das2020detecting}. More recently, \citet{pramanick2021detecting} firstly defined the harmful meme concept and demonstrated its dependence on contextual factors. The complex nature of memes, which often rely on multiple modalities, makes them challenging and struggle to yield good performance only using unimodal detection methods~\cite{simonyan2014very, he2016deep, devlin2018bert}. Therefore, recent studies in this area attempted to apply multimodal approaches on the harmful meme detection task.

Previous studies have employed classical two-stream models that integrate text and vision features, which are learned from text and image encoders, using attention-based mechanisms and multimodal fusion techniques for classifying harmful memes~\cite{kiela2019supervised, kiela2020hateful, suryawanshi2020multimodal}. Another branch was to fine-tune pre-trained multimodal models specifically for the task~\cite{lippe2020multimodal, muennighoff2020vilio, velioglu2020detecting, hee2022explaining}. Recent related efforts have also sought to explore the use of data augmentation techniques~\cite{zhou2021multimodal, zhu2022multimodal}, ensemble methods~\cite{zhu2020enhance, velioglu2020detecting, sandulescu2020detecting} and harmful target disentanglement~\cite{lee2021disentangling}.  More recently, \citet{pramanick2021momenta} proposed a multimodal framework by using global and local perspectives to detect harmful memes which achieves state-of-the-art performances. %to a state-of-the-art multimodal framework was designed by  to use global and local perspectives to detect harmful memes. 
A follow-up prompt-based approach~\cite{cao2023prompting} attempted to concatenate the meme text and extracted image captions to fine-tune masked language models~\cite{liu2019roberta} for harmful meme detection. However, existing solutions only capture the superficial signals of different modalities in memes in an end-to-end manner, which largely ignore explicit deductive reasoning to guide the model for understanding background knowledge about the complex and diverse relations between the visual and textual elements. 
% \lc{it would be better to high-level summarize these existing works, instead of just listing A xxxx, B xxxx, and C xxxx.}
% meanwhile, only using the captions as opposed to original vision features may suffer from a lack of mutual synergy in the representation space of different modalities due to the inductive bias of possible information loss in the captioning process.
\subsection{Large Language Models}
Recently, LLMs have demonstrated remarkable capability in complex reasoning~\cite{brown2020language, thoppilan2022lamda, rae2021scaling, chowdhery2022palm}, such as generating intermediate inference procedures before the final output% reasoning steps before arriving at the final output
~\cite{nye2021show, weichain, kojima2022large, zhang2022automatic}. Unfortunately, the large size of LLMs restricts their deployment on detecting harmful memes with %detection task of 
different modalities, regardless of how they are enhanced with strategetic text prompting. Knowledge distillation has been successfully used to transfer knowledge from larger, more competent teacher models into smaller student models affordable for practical applications~\cite{buciluǎ2006model, hinton2015distilling, beyer2022knowledge}. However, existing researches on knowledge distillation from LLMs~\cite{wang2022pinto, ho2022large, magister2022teaching} only consider the language modality, they are not suitable for harmful meme detection because harmful memes can convey holistic synergy information through multimodal features.
%Despite the flourishing of the research in knowledge distillation from LLMs~\cite{wang2022pinto, ho2022large, magister2022teaching} that are largely isolated in the language modality, there is little attention being put on harmful meme detection, of conveying holistic synergy information through multimodal features. 
In this work, we conduct abductive reasoning with LLMs, which further advocates a multimodal reasoning paradigm to fine-tune smaller language models (LMs) for harmful meme detection.

% Despite the flourishing of the research in CoT methods~\cite{wang2022rationale, zhou2022least, lu2022dynamic, fu2022complexity} that are largely isolated in the language modality, there is little attention being put on the harmful memes detection task. In this paper, we elicit abductive reasoning with LLMs, which further advocates a multimodal CoT reasoning paradigm to fine-tune smaller language models (LMs) for harmful memes detection.

\begin{figure*}[t]
\centering
\scalebox{0.8}{\includegraphics[width=20cm]{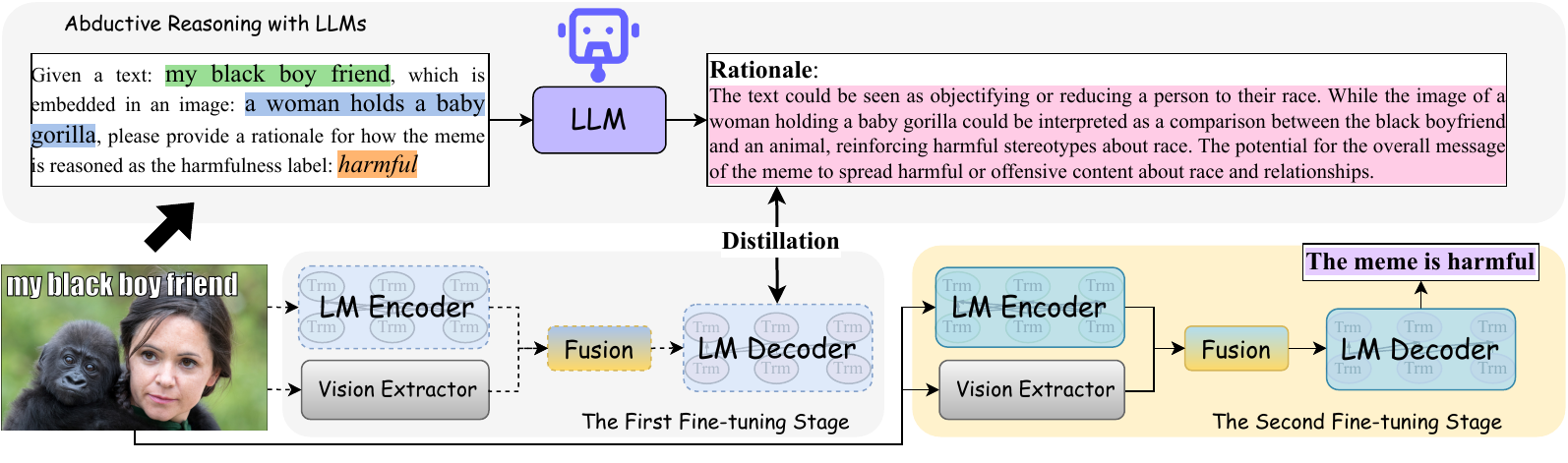}}
\vspace{-0.5cm}
\caption{The overall pipeline of our method. We first conduct abductive reasoning with LLMs to extract harmfulness rationales (\textcolor{magenta}{pink}) by the prompt consisting of the meme text (\textcolor{green(ncs)}{green}), the image caption (\textcolor{frenchblue}{blue}), and the label (\textcolor{orange}{orange}). We then use the generated rationales to train small task-specific models with multimodal inputs as the first fine-tuning stage and feed the same inputs to the updated model for harmfulness inference as the second fine-tuning stage.}
\label{fig:method}
\end{figure*}

% \section{Problem Statement}
% \lc{this section can be merged into the Approach section.}

% We define a harmful meme detection dataset for training as a set of memes $\mathbb{D}_s = \{M^s_1, M^s_2, ..., M^s_{|\mathbb{D}_s|}\}$, where each meme $M^s=\{y, \mathcal{I}, \mathcal{T}\}$ is a triplet representing an image $\mathcal{I}$ that is associated with a text $\mathcal{T}$, and its harmfulness label $y \in \{\text{harmful}, \text{harmless}\}$. And the dataset for testing is a set of memes $\mathbb{D}_t = \{M^t_1, M^t_2, ..., M^t_{|\mathbb{D}_t|}\}$ without labels, where each meme $M^t=\{\mathcal{I}', \mathcal{T}'\}$ is a tuple just representing an image $\mathcal{I}'$ and a text $\mathcal{T}'$.
% \lc{if we remove the last sentence in the next paragraph, you don't need D anymore. Here, you can only use M=(I, T, y).}

% In this work, to investigate multimodal reasoning distilled from large language models, we convert the harmful meme detection task into a natural language modeling paradigm, where our model takes the text $\mathcal{T}$ and image $\mathcal{I}$ as the input and generates a text sequence that consists of the harmfulness label $y$.
% \textcolor{red}{So, this task could be formulated as a supervised classification problem that trains a model $f(\cdot)$ from the labeled memes in the training set to detect the unlabeled meme in the test set, that is, $f(M^t| \mathbb{D}_s) \rightarrow y'$, where $y' \in \{\text{harmful}, \text{harmless}\}$. 
% }\lc{from my personal prospectively, the last sentence is useless.}

\section{Our Approach}
\paragraph{Problem Statement} We define a harmful meme detection dataset as a set of memes where each meme $M=\{y, \mathcal{I}, \mathcal{T}\}$ is a triplet representing an image $\mathcal{I}$ that is associated with a text $\mathcal{T}$, and a ground-truth harmfulness label $y \in \{\texttt{harmful}, \texttt{harmless}\}$. In this work, to investigate multimodal reasoning distilled from LLMs, we convert the harmful meme detection task into a natural language generation paradigm, where our model takes the text $\mathcal{T}$ and image $\mathcal{I}$ as the input and generates a text sequence that contains the label $y$ to clearly express whether the meme is harmful. %consists of the harmfulness label $y$.

Our core idea is to reason and evolve with the cognition-level rationale beyond the recognition-level perception~\cite{davis2015commonsense} by capturing the inter-relationship between visual and textual elements in memes. The overview of our framework is shown in Figure~\ref{fig:method}, which consists of abductive reasoning with LLMs (see Sec.~\ref{sec:3.1}) and two training stages, 
\ie, reasoning distillation (see Sec.~\ref{sec:3.2}) and harmfulness inference (see Sec.~\ref{sec:3.3}).

%without needing to check new rationales’ fluency and correctness
\subsection{Abductive Reasoning with LLMs} \label{sec:3.1} % to solve the problem of lacking gold rationale   1: image description  2: meme text, image description and harmfulness label -> rationale
% Recent studies observe one intriguing emerging property of LLMs: their ability to generate intermediate rationales that support their predictions~\cite{weichain, bang2023multitask}.
% \textcolor{red}{Previous literature revealed the ability of LLMs to generate intermediate rationales for improving classification tasks%support their predictions
% ~\cite{weichain, bang2023multitask}. While the studies only focused on how to elicit such reasoning capability from LLMs~\cite{nye2021show, weichain, kojima2022large}}\lc{this part is not suitable for the Approach section, it would be better to move it to the Related Work section.}, 
In this paper, we propose to utilize abductive reasoning with multimodal inputs to train smaller downstream models. %use them to assist smaller downstream model training, by abductive reasoning with multimodal inputs. The core idea is to view LLMs as teachers that can reason: 
LLMs can produce natural language rationales unveiling the implicit meaning beneath the surface of the memes to justify the reason why the meme is harmful or not. %the given harmfulness labels in training data. 
This shares a similar intuition as heuristic teaching~\cite{pintrich2002motivation} where a teacher who has rich experience and knowledge can impart to students the correct way of thinking and reasoning based on questions with corresponding answers. The students then learn how to deduce their own ways to the correct answers from questions accordingly. Thus we aim to activate explicit reasoning knowledge in LLMs as a teacher model, \eg, contextual and cultural information related to memes, to guide our model to strengthen harmfulness prediction.

Given a meme sample $M=\{y, \mathcal{I}, \mathcal{T}\}$ from the training data, to prompt large language models in uniform language modality, we first extract the text caption $\tilde{\mathcal{I}}$ of the image $\mathcal{I}$ by off-the-shelf captioning models~\cite{mokady2021clipcap}. Then we curate a template $p$ that consists of a triplet $\{y, \tilde{\mathcal{I}}, \mathcal{T}\}$ as observed attributes, to prompt the LLMs to generate a rationale $r$ that elicits the reasoning knowledge about how to infer the harmfulness label $y$ based on the interplay of the meme text $\mathcal{T}$ and the image caption $\tilde{\mathcal{I}}$ as illustrated in Figure~\ref{fig:method}. Specifically, we design $p$ as:

``\textit{Given a Text: [$\mathcal{T}$], which is embedded in an Image: [$\tilde{\mathcal{I}}$]; and a harmfulness label [$y$], please give me a streamlined rationale associated with the meme, without explicitly indicating the label, for how it is reasoned as [$y$].}''

As we clarify the ground-truth harmfulness label in the observed attributes of the prompt, the hallucination issue~\cite{bang2023multitask} of LLMs could be effectively alleviated. Because the rich contextual background knowledge could be activated by abductive reasoning based on the ground truth and invalid rationales are naturally filtered out.% during text generation. %constrained to avoid invalid rationales for mistake prediction, and the rich contextual background knowledge could be activated by abductive reasoning conducted on the ground truth.

\subsection{Reasoning Distillation with Small LMs} \label{sec:3.2}
Since we utilize image captions to represent the meme images, we could perform abductive reasoning with large language models pre-trained with language modality. 
However, only using the captions as opposed to original vision features may suffer from a lack of mutual synergy in the representation space of different modalities in memes due to the inductive bias of possible information loss in the captioning process. On the other hand, LLMs can be used to conduct abductive reasoning only for the training data whose harmfulness label is given in prior but is challenging to be fine-tuned for this task due to the huge amount of model parameters.
%serve testing in this multimodal task due to the large-scale model size. 
To facilitate the interactions between the meme text and the image, we propose to fine-tune a smaller language model for the harmful meme detection task, which allows flexibility in adjusting model architectures to incorporate multimodal features and is more lightweight for task-specific fine-tuning.  %The model training consists of two fine-tuning stages: 1) Reasoning Distillation and 2) Harmfulness Inference. In this section we would introduce the first stage and the second would be depicted in Sec.~\ref{harmfulness_inference}.
% \textcolor{red}{I will move the fine-tuning part to Sec 4.3}
% \lc{is there any other advantages for this small LM design?}

In this section, we train a small language model 
%In the first fine-tuning stage, the language model is trained 
as a student model distilled from the LLMs with multimodal reasoning knowledge. Specifically, we leverage generated rationales from LLMs as informative supervision, to fine-tune a smaller pre-trained language model to excavate the rich inter-relationship between language and vision modalities of memes.

\paragraph{Encoding} For a meme sample $M$ from the training data, we first encode the text $\mathcal{T}$ and the image $\mathcal{I}$ to obtain their embedding vectors as follows:
{
\begin{equation}
\begin{aligned}
    H_{\mathcal{T}}^0 &= \operatorname{TE}(\mathcal{T}),\\
    H_{\mathcal{I}} &= \operatorname{VE}(\mathcal{I}),
\end{aligned}
\end{equation}}
where $\operatorname{TE}(\cdot)$ denotes the text embedding layer of the LM Encoder. And $H_{\mathcal{T}}^0 \in \mathbb{R}^{m\times d}$ is the token embeddings in the Transformer encoder~\cite{vaswani2017attention} where $m$ is the text token length and $d$ is the dimension of the hidden states. $\operatorname{VE}(\cdot)$ is the Vision Extractor implemented as frozen pre-trained vision Transformers~\cite{radford2021learning} to fetch the patch-level features of the image with $n$ patches, which is projected into the visual representations $H_{\mathcal{I}} \in \mathbb{R}^{n\times d}$. Next, to support semantic alignment between the text and the image for better context understanding, we exploit a cross-attention mechanism~\cite{Luo2022ITuningTF} for multimodal fusion of the textual and visual information: 
{\setlength{\abovedisplayskip}{0.1cm}
\setlength{\belowdisplayskip}{0.1cm}
\begin{equation}
\begin{aligned}
    Q_{\mathcal{T}} &= W_{Q}^iH_{\mathcal{T}}^i + b_Q^i,\\
    K_{\mathcal{I}} &= W_{K}^iH_{\mathcal{I}} + b_K^i,\\
    V_{\mathcal{I}} &= W_{V}^iH_{\mathcal{I}} + b_V^i,\\  H_{\mathcal{I}}^i&=\operatorname{softmax}\left(\frac{Q_{\mathcal{T}}K^{\top}_{\mathcal{I}}}{\sqrt{d_k}}\right)V_{\mathcal{I}},
\end{aligned}
\end{equation}}
where $H_{\mathcal{T}}^i$ is the input hidden states of each LM Encoder layer and $H_{\mathcal{I}}^i$ is the attended visual features. Then we can fuse $H_{\mathcal{I}}^i$ with $H_{\mathcal{T}}^i$ to attain the interplay representations for a meme:
{\setlength{\abovedisplayskip}{0.1cm}
\setlength{\belowdisplayskip}{0.1cm}
\begin{equation}
    \begin{aligned}
     H_{\mathcal{T}}^{i+1} &= \operatorname{LME}^i\left(H_{\mathcal{T}}^{i}\right)
     + W_{O}^iH_{\mathcal{I}}^i + b_O^i,
    \end{aligned}
\end{equation}}
where $\operatorname{LME}^i(\cdot)$ is the $i$-th layer of the LM Encoder, $W_{*}^i$ denotes the linear projection, $b_{*}^{i}$ is the bias, and $\hat{H}=H_{\mathcal{T}}^{L}$ is the final interplay representations after going through an L-layer LM Encoder fused with the visual features.

\paragraph{Decoding} Finally, we feed the interplay representations $\hat{H}  \in \mathbb{R}^{m \times d}$ into the LM Decoder, implemented as a Transformer-based decoder, to generate the reasonable rationale. Overall, the smaller language model $f$ is trained by minimizing the following distillation loss:
{\setlength{\abovedisplayskip}{0.1cm}
\setlength{\belowdisplayskip}{0.1cm}
\begin{equation}
    \mathcal{L}_{distill} = \operatorname{CE}\left(f(\mathcal{I}, \mathcal{T}), r \right),
\end{equation}}
where $\operatorname{CE}(\cdot)$ denotes the cross-entropy loss~\cite{sutskever2014sequence} between the predicted text and the target rationale $r$ generated by LLMs. In this way, multimodal reasoning knowledge about the meme could be explicitly distilled from LLMs and injected into the smaller language model specific to harmful meme detection.

\subsection{Harmfulness Inference} \label{harmfulness_inference} \label{sec:3.3}
During the first fine-tuning stage, we conducted explicit deductive reasoning to empower our model with the capability of multimodal reasoning distilled from LLMs. As the goal of this task is to determine whether the meme is harmful or not, we conduct the second fine-tuning stage for Harmfulness Inference, which shares the same model architecture, parameters, and encoding procedure as Sec.~\ref{sec:3.2} %the Reasoning Distillation stage 
but differs in the decoding output. %target of the decoding procedure. 
To make the output consistent with harmfulness prediction, the smaller model $f$ is further trained by minimizing the following inference loss:
{\setlength{\abovedisplayskip}{0.1cm}
\setlength{\belowdisplayskip}{0.1cm}
\begin{equation}
    \mathcal{L}_{infer} = \operatorname{CE}\left(f(\mathcal{I}, \mathcal{T}), y \right),
\end{equation}}
where the cross-entropy loss is computed between the generated text and ground-truth harmfulness label $y$. With the generative objective~\cite{raffel2020exploring} adapted to the previous Reasoning Distillation stage, the prior reasoning knowledge absorbed in Reasoning Distillation could be well induced for Harmfulness Inference.

\paragraph{Model Training} The model training consists of two fine-tuning stages: 1) Reasoning Distillation and 2) Harmfulness Inference, where Reasoning Distillation is the predecessor fine-tuning phase of Harmfulness Inference. Note that for model testing, we directly input the test sample into our fine-tuned language model to predict the meme harmfulness. 

% \begin{table}[t]
% \centering
% \resizebox{0.45\textwidth}{!}{\begin{tabular}{@{}ccccc@{}}
% \toprule
% \multirow{2}{*}{Datasets} & \multicolumn{2}{c}{Train} & \multicolumn{2}{c}{Test} \\
%                           & \#harmful    & \#harmless   & \#harmful   & \#harmless   \\ \midrule
% Harm-C                    & 1064        & 1949        & 124        & 230         \\
% Harm-P                    & 1486        & 1534        & 173        & 182         \\
% FHM                       & 3050        & 5450        & 250        & 250         \\ \bottomrule
% \end{tabular}}
% \caption{Statistics of Datasets.}
% \label{tab:statistics}
% \end{table}

\begin{table*}[t]
    \centering
\resizebox{0.95\textwidth}{!}{\begin{tabular}{@{}l||cc|cc|cc@{}}
\toprule
Dataset         & \multicolumn{2}{c|}{Harm-C}                  & \multicolumn{2}{c|}{Harm-P}                        & \multicolumn{2}{c}{FHM}                     \\ \midrule
Model           & Accuracy                 & Macro-$\emph{F}_1$                & Accuracy                 & Macro-$\emph{F}_1$                     & Accuracy                 & Macro-$\emph{F}_1$               \\ \midrule \midrule
Text BERT~\cite{devlin2018bert}       & 70.17                & 66.25                 & 80.12                & \multicolumn{1}{c|}{78.35} & 57.12                &  41.52                    \\
Image-Region~\cite{he2016deep}    & 68.74                & 62.97                 & 73.14                & \multicolumn{1}{c|}{72.77} & 52.34                & 34.19                \\ \midrule
Late Fusion~\cite{pramanick2021detecting}     & 73.24                & 70.25                 & 78.26                & \multicolumn{1}{c|}{78.50} & 59.14                &  44.81                    \\
MMBT~\cite{kiela2019supervised}            & 73.48                & 67.12                 & 82.54                & \multicolumn{1}{c|}{80.23} & 65.06                &  61.93                    \\
VisualBERT COCO~\cite{li2019visualbert} & 81.36                & 80.13                 & 86.80                & \multicolumn{1}{c|}{86.07} & 61.48                & 47.26                \\
ViLBERT CC~\cite{lu2019vilbert}      & 78.70                & 78.09                 & 87.25                & \multicolumn{1}{c|}{86.03} & 64.70                &  55.78                    \\
MOMENTA~\cite{pramanick2021momenta}         & 83.82                & \underline{82.80}                 & \textbf{89.84}                & \multicolumn{1}{c|}{\underline{88.26}} & 61.34                &   57.45                   \\
MaskPrompt~\cite{cao2023prompting}      & \underline{84.47}                & 81.51                 & 88.17                & \multicolumn{1}{c|}{87.09} & \underline{72.98}                & \underline{65.24}                \\ \midrule 
\textsc{Mr.}\textsc{Harm}            & \multicolumn{1}{c}{\textbf{86.16}} & \multicolumn{1}{c|}{\textbf{85.43}} & \multicolumn{1}{c}{\underline{89.58}} & \multicolumn{1}{c|}{\textbf{89.57}} & \multicolumn{1}{c}{\textbf{75.40}} & \textbf{75.10} \\\bottomrule
\end{tabular}}
    \caption{Harmful meme detection results on three datasets. The accuracy and macro-averaged F1 score (\%) are reported as the metrics. The best and second results are in bold and underlined.}
    \label{tab:main_results}
    \vspace{-0.4cm}
\end{table*}

\section{Experiments}
\subsection{Experimental Setup}
\paragraph{Datasets} We use three publicly available meme datasets for evaluation: (1) Harm-C~\cite{pramanick2021detecting}, (2) Harm-P~\cite{pramanick2021momenta}, and (3) FHM~\cite{kiela2020hateful}. Harm-C and Harm-P consist of memes related to COVID-19 and US politics, respectively. FHM was released by Facebook as part of a challenge to crowd-source multimodal harmful meme detection in hate speech solutions. Different from FHM that each meme was labeled as \textit{harmful} or \textit{harmless}, Harm-C and Harm-P were originally labeled with three classes: \textit{very harmful}, \textit{partially harmful}, and \textit{harmless}. For a fair comparison, we merge the \textit{very harmful} and \textit{partially harmful} memes into \textit{harmful} ones, following the evaluation setting of recent work~\cite{pramanick2021momenta, cao2023prompting}. %Thus, we have a more practical classification, \ie, \textit{harmful} and \textit{harmless}. 
We provide statistics of the three datasets in the Appendix.
% The statistics of the three datasets are shown in Table~\ref{tab:statistics}.

% We adopt the evaluation metrics Area Under the Receiver Operating Characteristic curve (AUROC) and Accuracy (Acc). In order to report more reliable results, we measure the average performance of models under ten random seeds. All models use the same set of random seeds.

\paragraph{Baselines} We compare \textsc{Mr.}\textsc{Harm} with several state-of-the-art harmful meme detection systems: 1) \textbf{Text BERT}~\cite{devlin2018bert}; 2) \textbf{Image-Region}~\cite{ren2016faster, he2016deep}; 3) \textbf{Late Fusion}~\cite{pramanick2021detecting}; 4) \textbf{MMBT}~\cite{kiela2019supervised}; 5) \textbf{VisualBERT COCO}~\cite{li2019visualbert, lin2014microsoft}; 6) \textbf{ViLBERT CC}~\cite{lu2019vilbert, sharma2018conceptual}; 7) \textbf{MOMENTA}~\cite{pramanick2021momenta}; 8) \textbf{MaskPrompt}~\cite{cao2023prompting}. We use the accuracy and macro-averaged F1 score as the evaluation metrics. More implementation details and baseline descriptions are provided in Appendix.

\begin{table*}[t]
    \centering
\resizebox{0.95\textwidth}{!}{\begin{tabular}{@{}l||cc|cc|cc@{}}
\toprule
Dataset         & \multicolumn{2}{c|}{Harm-C}                  & \multicolumn{2}{c|}{Harm-P}                        & \multicolumn{2}{c}{FHM}                     \\ \midrule
Model           & Accuracy                 & Macro-$\emph{F}_1$                & Accuracy                 & Macro-$\emph{F}_1$                     & Accuracy                 & Macro-$\emph{F}_1$               \\ \midrule \midrule
\textsc{Mr.}\textsc{Harm}            & \multicolumn{1}{c}{86.16} & \multicolumn{1}{c|}{85.43} & \multicolumn{1}{c}{89.58} & \multicolumn{1}{c|}{89.57} & \multicolumn{1}{c}{75.40} & {75.10} \\
\hspace{0.5cm} w/o Reasoning Distillation & \multicolumn{1}{c}{83.33} & \multicolumn{1}{c|}{81.44} & \multicolumn{1}{c}{88.17} & \multicolumn{1}{c|}{88.17} & \multicolumn{1}{c}{73.60} & {73.41} \\
\hspace{0.5cm} w/o Visual Features & \multicolumn{1}{c}{82.48} & \multicolumn{1}{c|}{80.30} & \multicolumn{1}{c}{87.04} & \multicolumn{1}{c|}{87.03} & \multicolumn{1}{c}{58.80} & {57.01} \\
\hspace{0.5cm} w/o Multimodal Fusion & \multicolumn{1}{c}{79.38} & \multicolumn{1}{c|}{75.36} & \multicolumn{1}{c}{87.46} & \multicolumn{1}{c|}{87.45} & \multicolumn{1}{c}{74.40} & {74.25} \\
\hspace{0.5cm} w/o Two-stage Training & \multicolumn{1}{c}{83.05} & \multicolumn{1}{c|}{81.45} & \multicolumn{1}{c}{63.32} & \multicolumn{1}{c|}{63.32} & \multicolumn{1}{c}{67.40} & {65.77}\\
\hspace{0.5cm} w/o Fine-tuning Small LMs & \multicolumn{1}{c}{71.75} & \multicolumn{1}{c|}{66.86} & \multicolumn{1}{c}{61.13} & \multicolumn{1}{c|}{60.27}      & \multicolumn{1}{c}{60.00} & {57.72} \\ 
\bottomrule
\end{tabular}}
\vspace{-0.1cm}
    \caption{Ablation studies on our proposed framework.}
    \label{tab:ablation}
    \vspace{-0.3cm}
\end{table*}

\subsection{Harmful Meme Detection Performance}
Table~\ref{tab:main_results} shows the performance of our proposed method versus all the compared methods on the Harm-C, Harm-P and FHM datasets. It is observed that 1) The performance of the baselines in the first group is obviously poor due to only unimodal features like text-only or image-only being captured. In comparison, the other baselines exploit the multimodal features from both the text and image in memes. 2) The multimodal models in the second group outperform the unimodal ones. The early-fusion models with multimodal pre-training (\ie, VisualBERT COCO and ViLBERT CC) outperform that of the simple fusion with unimodal pre-training (\ie, Late Fusion and MMBT) on Harm-C/P datasets, while MOMENTA performs best in the second group by considering global and local information of memes. 3) However, as the images in FHM dataset are more informative and high-quality, MaskPrompt yields the best performance among all the baselines by %others on FHM dataset where the image is more informative and high-quality than that in Harm-C/P by 
incorporating additional extracted entities and demographic information of the image into the masked language models, besides just captioning the image into the prompt.

Our proposed \textsc{Mr.Harm} improves over the best baselines by %achieves superior performance among all the baselines with 
2.63\%, 1.31\%, and 9.86\% in terms of Macro-F1 score on Harm-C, Harm-P, and FHM datasets, respectively. We observe that 1) the improvement on the Harm-P dataset is relatively milder than that on the other two datasets. Meanwhile, all the baselines just have tiny differences among their performances on Harm-P. We speculate the reason falls into the smaller dataset scale of Harm-P which only 
%that the reason is that the models more readily comprehend Harm-P data since it has the smallest data volume and only
contains politics-related harmful memes. 2) A similar trend can also be observed in Harm-C and FHM datasets: the more challenging the dataset is, the greater performance improvement \textsc{Mr.Harm} achieves. Our model performs flexibly and stably across all datasets with its keen judgment on harmful memes. This is because all the baselines are only designed at the recognition level, but \textsc{Mr.Harm} is further empowered with multimodal reasoning knowledge distilled from LLMs to unearth harmful content from the seemly uncorrelated text and image modalities of memes.
%greater the improvement our model demonstrates in performance, 
%reflecting its keen judgment on harmful memes and indicating our model's flexibility and stability on different datasets. Different from the aforementioned baselines only at the recognition level, our model is empowered with multimodal reasoning knowledge distilled from large language models to unearth harmful content from the seemingly unrelated image and text modalities of memes.

\subsection{Ablative Study}
We perform ablative studies on several variants of \textsc{Mr.}\textsc{Harm}: 1) \textit{w/o Reasoning Distillation}: Simply fine-tune the smaller language models in the stage of Harmfulness Inference without the stage of Reasoning Distillation based on LLMs; %that distills reasoning knowledge from LLMs; 
2) \textit{w/o Visual Features}: Discard the features from the meme image while keeping those from the meme text; 3) \textit{w/o Multimodal Fusion}: Instead of the fusion mechanism on the multimodal features in our language model, we only append the lingual features from image captioning together with the meme text during encoding; 4) \textit{w/o Two-stage Training}: Concatenate the rationales generated from LLMs and golden harmfulness label as the target for model training, to replace the two-stage training paradigm; 5) \textit{w/o Fine-tuning Small LMs}: Directly prompt the representative large language model ChatGPT based on InstructGPT~\cite{ouyang2022training} for harmful meme detection.

As demonstrated in Table~\ref{tab:ablation}, the ablative models suffer different degrees of performance degradation, indicating the effectiveness of our proposed components for harmful meme detection with multimodal reasoning distilled from LLMs. Specifically, the performance of \textsc{Mr.}\textsc{Harm} significantly decreases in the `\textit{w/o Reasoning Distillation}' setting due to the lack of multimodal reasoning knowledge transferred from LLMs about the seemly uncorrelated modalities in memes. The `\textit{w/o Visual Features}' setting also achieves worse performance than \textsc{Mr.}\textsc{Harm}, suggesting that the visual representations are complementary to the meme text for harm-indicative pattern extraction in the language model. \textsc{Mr.}\textsc{Harm} makes improvements over `\textit{w/o Multimodal Fusion}', which implies the promoting role of our fusion mechanism that incorporates original vision features into the language model, hardly compromised when there could be severe information loss in the captioning process. Moreover, the `\textit{w/o Two-stage Training}' setting leads to large-margin performance degradation, which verifies the effectiveness of our two-stage training paradigm. This is because this setting causes mutual interference between intermediate reasoning and final prediction, which affects the convergence effect of harmfulness inference and damages the model's performance and stability. Compared with \textsc{Mr.}\textsc{Harm}, the performance of `\textit{w/o Fine-tuning Small LMs}' also significantly decreases, highlighting the importance of abductive reasoning with LLMs to alleviate the hallucination issue during deductive reasoning for harmfulness prediction.

\begin{figure}[t!]
\centering
\scalebox{0.37}{\includegraphics[width=20cm]{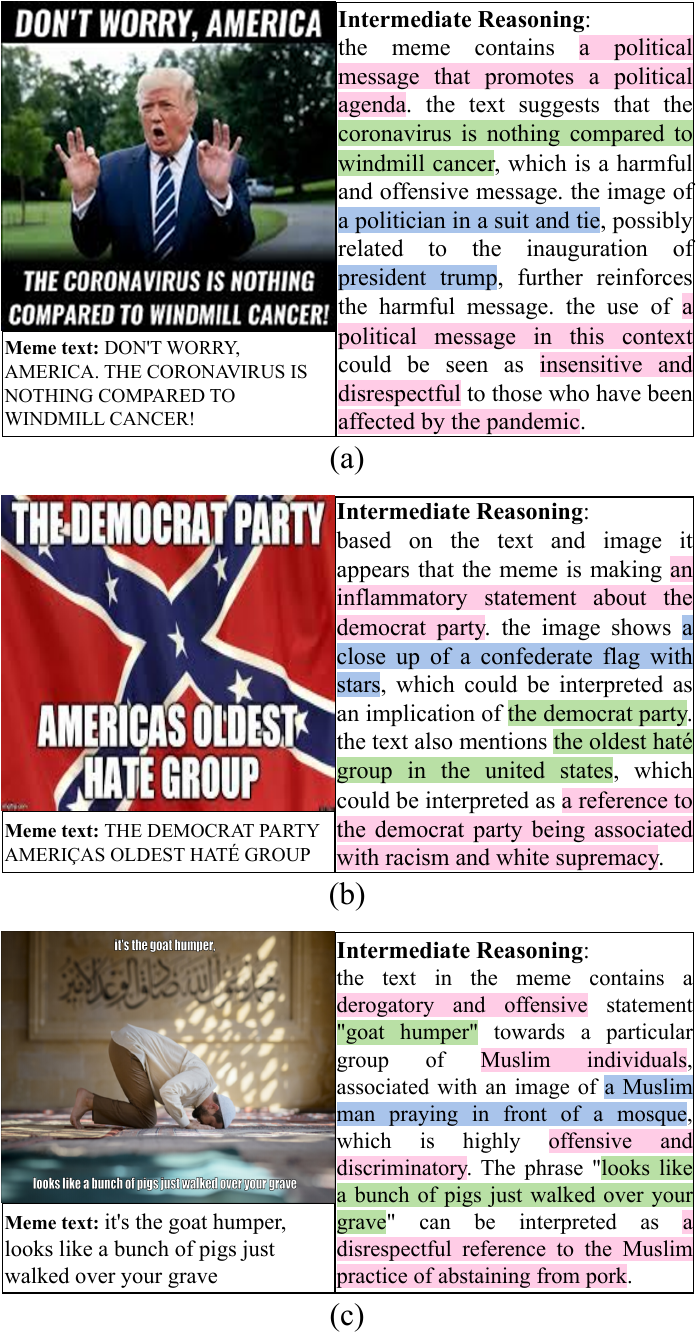}}
\vspace{-0.2cm}
\caption{Examples of correctly predicted harmful memes in (a) Harm-C, (b) Harm-P, and (c) FHM dataset.}
\label{fig:case}
\vspace{-0.1cm}
\end{figure}

% {
% % \setlength{\abovecaptionskip}{-0.1cm}
% % \setlength{\belowcaptionskip}{-0.1cm}
% \begin{figure*}[t]
% \centering
% \subfigure[Harm-C]{
% \begin{minipage}[t]{0.33\linewidth}
% \centering
% \scalebox{0.85}{\includegraphics[width=6cm]{baseline_comparision_HarmC_.pdf}}
% %\caption{fig1}
% \label{fig:pp_harmc}
% \end{minipage}%
% }%
% \subfigure[Harm-P]{
% \begin{minipage}[t]{0.33\linewidth}
% \centering
% \scalebox{0.85}{\includegraphics[width=6cm]{baseline_comparision_HarmP_.pdf}}
% %\caption{fig2}
% \label{fig:pp_harmp}
% \end{minipage}%
% }%
% \subfigure[FHM]{
% \begin{minipage}[t]{0.33\linewidth}
% \centering
% \scalebox{0.85}{\includegraphics[width=6cm]{baseline_comparision_FHM_.pdf}}
% %\caption{fig2}
% \label{fig:pp_FHM}
% \end{minipage}%
% }%
% \centering
% \vspace{-0.5cm}
% \caption{The performance of our \textsc{Mr.}\textsc{Harm} and other multimodal baselines with respect to the parameter size.}
% \label{fig:performance_param}
% \vspace{-0.2cm}
% \end{figure*}}

\subsection{Cognition-view Reasoning Analysis}
Note that our smaller language model is explicitly trained in the Reasoning Distillation stage for rationale generation to distill multimodal reasoning knowledge from LLMs. Although intermediate reasoning is not the final target output for harmful meme detection, after the first fine-tuning stage, we elicit reasonable thoughts from our smaller language model with the test samples as input, to understand the cognition view of our proposed \textsc{Mr.}\textsc{Harm} on the test meme samples more transparently and intuitively, as exemplified in Figure~\ref{fig:case}. 

From the visualized intermediate reasoning, we observe that 1) our model could understand the multimodal information related to the meme text (in green) and image (in blue) with commonsense knowledge. For example, in Figure~\ref{fig:case}(a), the recognized ``politician” in the image could be related to ``president trump”, which could be linked to the ``AMERICA” in the text; in Figure~\ref{fig:case}(b), the recognized ``flag” in the image could be cognized to satire ``the democrat party” in the text; and in terms of Figure~\ref{fig:case}(c), the ``goat humper” and ``pigs” in the text could be associated with the attacks to ``a Muslim man” recognized in the image. 2) Furthermore, our model learns to cognize the interplay (in pink) of multimodal information with advanced reasoning. Benefitting from the rich multimodal understanding of the memes, the perpetuates harmful stereotypes could be reasoned over the context to the target like ``who affected by the pandemic” in Figure~\ref{fig:case}(a), ``the democrat party” in Figure~\ref{fig:case}(b), and ``the Muslim” in Figure~\ref{fig:case}(c). In this way, the rich correlation beneath the surface of the meme text and image could be excavated to facilitate harmfulness inference with better reasoning knowledge by harnessing advanced LLMs. Such readable pieces of rationales are also potentially valuable for aiding human checkers to verify the final answer predicted by our model. 

\begin{figure}[t]
\centering
\scalebox{0.37}{\includegraphics[width=20cm]{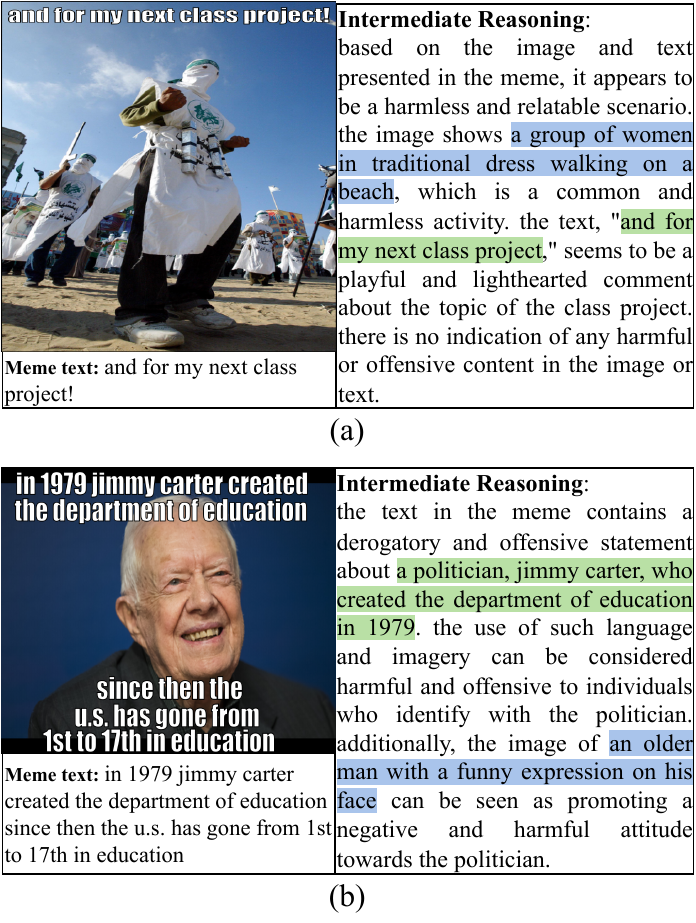}}
\vspace{-0.2cm}
\caption{Examples of wrongly predicted memes by our proposed framework with the ground truth (a) harmful and (b) harmless.}
\label{fig:error}
\vspace{-0.1cm}
\end{figure}

\subsection{Error Analysis}
To better understand the behavior of our model and facilitate future studies, we conduct an error analysis on the wrongly predicted memes by our proposed framework. We found that the major error exists in that our framework still cannot fully recognize the images that require rich background knowledge though we exploited the advanced cross-attention mechanism by incorporating visual features into the language model. Figure~\ref{fig:error} shows two examples of memes wrongly classified by \textsc{Mr.}\textsc{Harm}. For the harmful meme in Figure~\ref{fig:error}(a), the phrase ``and for my next class project!” suggests that the image is being used for an academic or educational purpose, which can be seen as glorifying or normalizing the behavior depicted in the image. The image features ``a group of Ku Klux Klan members walking on a beach”, which is a symbol of white supremacy and racism. The combination of the phrase in the text and the use of imagery associated with hate groups can contribute to the glorification of harmful behaviors and the perpetuation of negative stereotypes, which makes the meme harmful. However, due to the lack of related background knowledge about the Ku Klux Klan members and their wear, our framework cannot well recognize the image correctly during the original vision feature extraction, which leads to error propagation for wrongly concluding that the meme is harmless. Also, in terms of the harmless meme in Figure~\ref{fig:error}(b), the image that ``Jimmy Carter with a smile on his face” is mistakenly recognized as ``an older man with a \textit{funny} expression on his face”, furthermore, the model hallucinates that the meme text ``can be considered harmful and offensive to individuals who identify with the politician”, resulting in the wrong prediction that the meme is harmful. Therefore, it is possible to improve \textsc{Mr.}\textsc{Harm} by incorporating more informative vision features and improving language-vision interaction to be capable of understanding the images with more complex background knowledge.

\subsection{Discussion}
\begin{table}[t]\footnotesize
\centering
\resizebox{0.36\textwidth}{!}{\begin{tabular}{@{}cccc@{}}
\toprule
\multicolumn{1}{l}{Version} & Harm-C & Harm-P & FHM  \\ \midrule
Small                & 84.99   & 85.33   & 72.96 \\
Base                 & 85.43   & 89.57   & 75.10 \\
Large                & 84.02   & 90.14   & 77.80 \\ \bottomrule
\end{tabular}}
\vspace{-0.1cm}
\caption{Macro-averaged F1 score (\%) achieved with different versions of our fine-tuned LMs.}
\label{tab:T5}
\vspace{-0.4cm}
\end{table}

As our two-stage training paradigm requires distilling the reasoning knowledge and leveraging original vision features, we utilize the T5 encoder-decoder architecture~\cite{raffel2020exploring, chung2022scaling} to initialize our generative framework. To test the generality of the benefits of our approach to different versions of the backbone, we alter the underlying LMs to other variants in different sizes. As shown in Table~\ref{tab:T5}, one interesting phenomenon is that our model has already achieved outstanding performance on the three benchmarks with the \textit{Small} (about 60M parameters) or \textit{Base} ((about 220M parameters)) version as the backbone, which has a smaller size than the state-of-the-art baseline MaskPrompt (over 300M parameters). The \textit{Large} version of our backbone generally achieved better performance than the other two backbone versions because the larger the fine-tuned LMs, the more it alleviates the hallucination issue~\cite{ji2023survey}. 
% We further provide the comparison of performance on the three meme datasets with regard to the number of trainable parameters for \textsc{Mr.}\textsc{Harm} and the other multimodal baselines in Figure~\ref{fig:performance_param}. 
Overall, the results show that our framework does not rely excessively on the size of the backbone to improve performance and is generally effective with different versions of the backbone model.

\section{Conclusion and Future Work}
In this paper, we propose to capture implicit meaning that is not explicitly conveyed through the surface of the text and image in memes for harmful meme detection. We first conduct abductive reasoning with LLMs. Then we present a novel generative framework to distill multimodal reasoning knowledge from LLMs, which includes two training stages: 1) reasoning distillation and 2) harmfulness inference. Results on three meme benchmarks confirm the advantages of our proposed framework. For future work, since it is harder to judge the quality of the intermediate reasoning, where the evaluation is necessarily qualitative, we plan to do some sort of systematic study towards explainable harmful meme detection to claim explainability through a human subjects study for evaluation.

\section*{Limitations}
There are multiple ways to further improve this work:
\begin{itemize}
    \item Despite this work focusing on performance improvement of harmful meme detection, it is harder to judge the quality of the intermediate reasoning, where the evaluation is necessarily qualitative. Considering that our framework could generate readable snippets for cognition-view reasoning, we plan to do some sort of systematic study to claim explainability for the evaluation, which would be another more targeted research.
    \item New benchmarks to evaluate the reasoning ability of our framework are demanded. We are going to further exploit LLMs toward explainable harmful meme detection from the perspectives like dataset construction and automatic evaluation.
    \item We only use the textual prompt to conduct abductive reasoning with accessible LLMs pre-trained with the language modality. We would further update our framework by leveraging visual LLMs if accessible in the future to improve the visual feature extraction for exploring better multimodal reasoning knowledge distillation, and avoid several common deficiencies of existing language models including hallucination and limited generalization as much as possible.
\end{itemize}

\section*{Acknowledgements}
This work was partially supported by Hong Kong RGC ECS (Ref. 22200722) and National Natural Science Foundation of China Young Scientists Fund (No. 62206233).

\section*{Broader Impact}
The purpose of this work is to prevent the spread of harmful meme information and to ensure that people are not subjected to prejudice, racial and gender discrimination. Nevertheless, we are aware of the potential for malicious users to reverse-engineer and create memes that go undetected or misunderstood by \textsc{Mr.}\textsc{Harm}-trained AI systems. This is strongly discouraged and condemned. Intervention with human moderation would be required in order to ensure that this does not occur.

\bibliography{anthology,custom}
\bibliographystyle{acl_natbib}

\clearpage

\appendix

\section{Datasets}
\label{sec:appendix}
\begin{table}[t]
\centering
\resizebox{0.45\textwidth}{!}{\begin{tabular}{@{}ccccc@{}}
\toprule
\multirow{2}{*}{Datasets} & \multicolumn{2}{c}{Train} & \multicolumn{2}{c}{Test} \\
                          & \#harmful    & \#harmless   & \#harmful   & \#harmless   \\ \midrule
Harm-C                    & 1064        & 1949        & 124        & 230         \\
Harm-P                    & 1486        & 1534        & 173        & 182         \\
FHM                       & 3050        & 5450        & 250        & 250         \\ \bottomrule
\end{tabular}}
\caption{Statistics of Datasets.}
\label{tab:statistics}
\end{table}

The detailed statistics of the three datasets are shown in Table~\ref{tab:statistics}.

\section{Implementation Details}
To separate the text and image in the memes, we first in-paint the memes by combining MMOCR~\cite{kuang2021mmocr} with SAM~\cite{kirillov2023segment} to extract the text and pure image. Then during the captioning process, since the focus of this work is primarily on the multimodal reasoning for harmful meme detection from a fresh perspective on harnessing LLMs, we apply a pre-trained image captioning model ClipCap~\cite{mokady2021clipcap} used in recent work~\cite{cao2023prompting}, to generate textual descriptions about the dominant objects or events in the memes' image, which is utilized as the inputs into LLMs for abductive reasoning. To generate the rationale for each meme, we employed ChatGPT~\cite{ouyang2022training}, a widely used LLM developed by OpenAI, specifically utilizing the ``gpt-3.5-turbo'' version. To make our results reproducible, we set the temperature as 0 and the maximum length as 256.

For the system prompt to the ``gpt-3.5-turbo'' model, we design the message as:

``\textit{You have been specially designed to perform abductive reasoning for the harmful meme detection task. Your primary function is that, according to a harmfulness label about an image with a text embedded, please provide a streamlined rationale, without explicitly indicating the label, for how it is reasoned as the given harmfulness label. The image and the textual content in the meme are often uncorrelated, but its overall semantics is presented holistically. Thus it is important to note that you are prohibited from relying on your own imagination, as your goal is to provide the most accurate and reliable rationale possible so that people can infer the harmfulness according to your reasoning about the background context and relationship between the given text and image.}”. 

\begin{table}
    \centering
    \resizebox{0.45\textwidth}{!}{\begin{tabular}{lccc}
    \toprule
    Hyper-Parameter  &  Harm-C & Harm-P & FHM\\
    \midrule
    \midrule
    \multicolumn{4}{l}{\text{First-Stage}}\\
    \midrule
      epoch  & 10 & 10 & 10\\
      batch size & 32 & 32 & 32\\
      Learning Rate & 5e-5 & 5e-5 & 5e-5\\
      Warmup Step & 0.1 & 0.1 & 0.1\\
      Warmup Strategy & Linear & Linear & Linear\\
      Image Size & 224 & 224 & 224\\
    \midrule
    \multicolumn{4}{l}{\text{Second-Stage}}\\
    \midrule
      epoch  & 30 & 30 & 30\\
      batch size & 32 & 32 & 32\\
      Learning Rate & 5e-5 & 5e-4 & 1e-4\\
      Warmup Step & 0.1 & 0.1 & 0.1\\
      Warmup Strategy & Linear & Linear & Linear\\
      Image Size & 224 & 224 & 224\\
    \bottomrule
    \end{tabular}}
    \caption{Hyper-parameters.}
    \label{tab:hyper}
\end{table}

Moreover, to prompt the LLMs to generate reasonable rationales with the triplet $\{y, \tilde{\mathcal{I}}, \mathcal{T}\}$ as observed attributes, we design the template $p$ for the user prompt as: 

``\textit{Given a Text: [$\mathcal{T}$], which is embedded in an Image: [$\tilde{\mathcal{I}}$]; and a harmfulness label [$y$], please give me a streamlined rationale associated with the meme, without explicitly indicating the label, for how it is reasoned as [$y$].}”.

Our \textsc{Mr.}\textsc{Harm} model utilizes the T5 encoder-decoder architecture~\cite{raffel2020exploring, chung2022scaling} as its foundational framework, specifically utilizing the “flan-t5-base” version. For the extraction of image features, following previous work~\cite{pramanick2021momenta}, we adopted the state-of-the-art vision Transformer known as CLIP-ViT-B/32~\cite{radford2021learning}, and this module remains static throughout the training process. To effectively integrate the multi-modal information, we incorporated a simple one-head cross-attention mechanism in each layer of the T5 encoder. During the fusion process, the text features are utilized as the query, while the image features act as the key and value. It is noteworthy that these fusion modules were initialized randomly. For the fine-tuning phase, we provide a comprehensive list of the hyper-parameters in Table~\ref{tab:hyper}. Results are averaged over ten random runs. All experiments were conducted using a single V100 32GiB GPU. 

\begin{table*}[t] \footnotesize
    \centering
\resizebox{0.85\textwidth}{!}{\begin{tabular}{@{}l||cc|cc|cc@{}}
\toprule
Dataset         & \multicolumn{2}{c|}{Harm-C}                  & \multicolumn{2}{c|}{Harm-P}                        & \multicolumn{2}{c}{FHM}                     \\ \midrule
Version          & Accuracy                 & Macro-$\emph{F}_1$                & Accuracy                 & Macro-$\emph{F}_1$                     & Accuracy                 & Macro-$\emph{F}_1$               \\ \midrule 
           Small & 85.59 & 84.99 & 85.35 & 85.33 & 73.20 & 72.96\\
           Base & 86.16 & 85.43 & 89.58 & 89.57 & 75.40 & 75.10\\
           Large & 85.03 & 84.02 & 90.14 & 90.14 & 78.20 & 77.80\\
           \bottomrule
\end{tabular}}
    \caption{The detailed results with different sizes of our fine-tuned LMs.}
    \label{tab:size}
\end{table*}

\begin{table}[t]\Large
\centering
\resizebox{0.49\textwidth}{!}{\begin{tabular}{@{}lccc@{}}
\toprule
Models                & MOMENTA & MaskPrompt & \textsc{Mr.}\textsc{Harm} \\ \midrule
Multimodal Fusion & {\cmark} & \xmark     &  \cmark     \\
Prompt Tuning & \xmark & \cmark     & \cmark      \\
Explicit Reasoning  & \xmark   & \xmark     & \cmark        \\
Leveraging LLMs  & \xmark   & \xmark     & \cmark        \\\bottomrule
\end{tabular}}
\caption{Comparison of characteristics between our \textsc{Mr.}\textsc{Harm} with state-of-the-art models for harmful meme detection.}
\label{tab:comparison}
\end{table}

\begin{table*}[t]\footnotesize
    \centering
\resizebox{0.85\textwidth}{!}{\begin{tabular}{@{}l||cc|cc|cc@{}}
\toprule
Dataset         & \multicolumn{2}{c|}{Harm-C}                  & \multicolumn{2}{c|}{Harm-P}                        & \multicolumn{2}{c}{FHM}                     \\ \midrule
Model           & Accuracy                 & Macro-$\emph{F}_1$                & Accuracy                 & Macro-$\emph{F}_1$                     & Accuracy                 & Macro-$\emph{F}_1$               \\ \midrule \midrule
Explanation & \multicolumn{1}{c}{83.05} & \multicolumn{1}{c|}{81.45} & \multicolumn{1}{c}{63.32} & \multicolumn{1}{c|}{63.32} & \multicolumn{1}{c}{67.40} & {65.77}\\
Reasoning & \multicolumn{1}{c}{68.93} & \multicolumn{1}{c|}{56.19} & \multicolumn{1}{c}{56.90} & \multicolumn{1}{c|}{56.67} & \multicolumn{1}{c}{63.00} & {59.29} \\ 
\bottomrule
\end{tabular}}
    \caption{Effects of the one-stage training.}
    \label{tab:one-stage}
\end{table*}

\section{Baselines}
We compare our model \textsc{Mr.}\textsc{Harm} with several state-of-the-art harmful meme detection systems: 1) \textbf{Text BERT}: BERT~\cite{devlin2018bert} is utilized as the unomodal text-only model; 2) \textbf{Image-Region}: a unimodal visual-only model that processes meme images using Faster R-CNN~\cite{ren2016faster} with ResNet-152~\cite{he2016deep} to feed into a classification layer; 3) \textbf{Late Fusion}: a multimodal model uses the average prediction scores of BERT and ResNet-152 for harmful meme detection~\cite{pramanick2021detecting}; 4) \textbf{MMBT}: a multimodal Bi-Transformer~\cite{kiela2019supervised} that captures the intra-modal and inter-modal dynamics of the two modalities; 5) \textbf{VisualBERT COCO}: Visual BERT~\cite{li2019visualbert} pre-trained on the COCO dataset~\cite{lin2014microsoft}; 6) \textbf{ViLBERT CC}: Vision and Language BERT~\cite{lu2019vilbert} trained on an intermediate multimodal objective~\cite{sharma2018conceptual} for task-agnostic joint representations of image and text; 7) \textbf{MOMENTA}: a multimodal harmful meme detection system~\cite{pramanick2021momenta} that takes the global and local information in two modalities of memes into account; 8) \textbf{MaskPrompt}: a prompt learning approach~\cite{cao2023prompting} that converts harmful meme detection as a masked language modeling problem based on RoBERTa-large~\cite{liu2019roberta}. We use accuracy and macro-averaged F1 score as the evaluation metrics, where the macro-averaged F1 is the more important metric owing to the imbalanced class prevalence (see Table~\ref{tab:statistics}), to capture competitive performance beyond the majority class.

While LLMs offer strong zero/few-shot performance as shown in the `w/o Fine-tuning Small LMs' setting in Table~\ref{tab:ablation}, they are challenging to serve in practice that requires at least 350GB GPU memory using specialized infrastructure for a single 175 billion LLM. This work presents a novel paradigm to leverage the reasoning ability and rich background knowledge of LLMs for better harmful meme detection, but just need to fine-tune the small language model even with a smaller size than the state-of-the-art baseline. Table~\ref{tab:size} illustrates the detailed results on the three meme datasets with different versions of our fine-tuned backbone model. Table~\ref{tab:comparison} illustrates the comparison of characteristics between \textsc{Mr.}\textsc{Harm} and the state-of-the-art baselines like MOMENTA and MaskPrompt. 

\begin{figure}[]
\centering
\scalebox{0.6}{\includegraphics{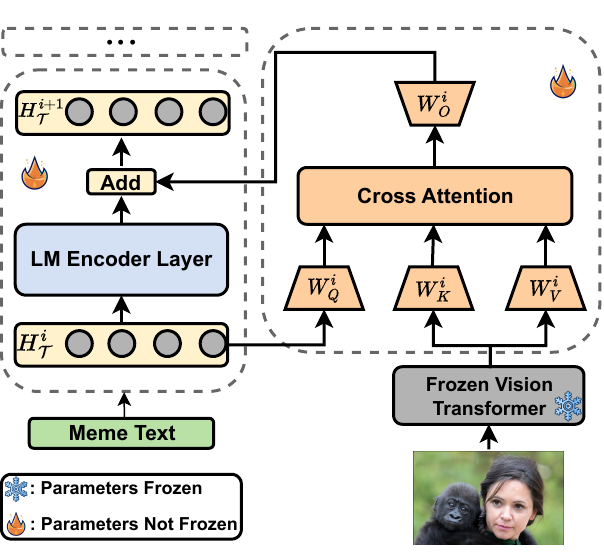}}
\caption{The details of our Multimodal Fusion module.}
\label{fig:mf}
\end{figure}

\section{Illustration of Multimodal Fusion}
Figure~\ref{fig:mf} illustrates the details of our multimodal fusion module in the encoding phase of \textsc{Mr.}\textsc{Harm}.

\section{Discussion about One-stage Training}

We further investigate the one-stage training to figure out the intrinsic property of the chain-of-thought reasoning. We compare the performance with two proposed variants for the one-stage training: 1) Explanation where the rationale is utilized for explaining the harmfulness inference; 2) Reasoning where harmfulness inference is conditioned to the rationale. As shown in Tabel~\ref{tab:one-stage}, the reasoning setting performs worse than the explanation setting with a large margin. We conjecture that this is because the reasoning setting in the one-stage training could lead to error propagation if our small language model generates hallucinated rationales that mislead the harmfulness inference, which however could be well avoided by the two-stage training paradigm. Meanwhile, as there exists mutual interference between rationale generation and harmfulness prediction, the explanation setting could give the harmfulness inference higher priority in the sequence generation so that it performs better than the reasoning setting. We argue that such a one-stage training paradigm could be improved in the future by applying a filtering mechanism, \eg, using only the effective chain-of-thought reasoning to infer the harmfulness of memes and get rid of irrelevant rationales. In summary, both settings in the one-stage training paradigm suffer different degrees of performance degradation, which reaffirms the necessity of our two-stage training paradigm.

\section{Future Work}
We will explore the following directions in the future:
\begin{itemize}
    \item Considering that our framework could generate readable snippets for cognition-view reasoning, we plan to do some sort of systematic study to claim explainability (possibly through a human subjects study) for the evaluation.
    \item In this work we target exploring the underlying reasoning process to empower the harmful meme detection model with the ability of explicit reasoning, to arrive at correct harmfulness predictions. We are going to further exploit LLMs toward explainable harmful meme detection from perspectives like dataset construction on social media with propagation structure~\cite{lin2021rumor, ma2020debunking,ma2020attention, lin2023zero}, automatic evaluation, and human evaluation.
    \item We would further update our framework by leveraging visual LLMs if accessible in the future to improve the visual feature extraction for better multimodal reasoning, and avoid several common deficiencies of existing language models including hallucination and limited generalization as much as possible.
\end{itemize}

% \begin{figure*}[t]
% \centering
% \scalebox{0.81}{\includegraphics[width=20cm]{llm_generated_rationale.pdf}}
% \caption{Examples of LLM-generated rationales for training meme samples labeled as harmful.}
% \label{fig:llm_generated}
% \end{figure*}

% \begin{figure*}[t]
% \centering
% \scalebox{0.81}{\includegraphics[width=20cm]{llm_generated_rationale_harmless.pdf}}
% \caption{Examples of LLM-generated rationales for training meme samples labeled as harmless.}
% \label{fig:llm_generated_}
% \end{figure*}

% \begin{figure*}[t]
% \centering
% \scalebox{0.81}{\includegraphics[width=20cm]{intermediate_reasoning.pdf}}
% \caption{Examples of intermediate reasoning for test meme samples labeled as harmful.}
% \label{fig:generated}
% \end{figure*}

% \begin{figure*}[t]
% \centering
% \scalebox{0.81}{\includegraphics[width=20cm]{intermediate_reasoning_harmless.pdf}}
% \caption{Examples of intermediate reasoning for test meme samples labeled as harmless.}
% \label{fig:generated_}
% \end{figure*}

\end{document}